\RequirePackage{fix-cm}
\documentclass{article}
\usepackage{graphicx}
\usepackage{hyperref}       
\usepackage{url}            
\usepackage{booktabs}       
\usepackage{amsfonts}       
\usepackage{nicefrac}       
\usepackage{microtype}      

\usepackage{wrapfig}
\usepackage{xcolor}
\usepackage{amssymb}

\begin{document}

\title{Improving Reinforcement Learning with Human Assistance: An Argument for Human Subject Studies with HIPPO Gym\thanks{This work has taken place in the Intelligent Robot Learning Lab at the University of Alberta, which is supported in part by research grants from the Alberta Machine Intelligence Institute (Amii), CIFAR, and NSERC.}
}




\author{
  Matthew E.~Taylor, Nicholas Nissen, Yuan Wang\\
  University of Alberta \&\\
  Alberta Machine Intelligence Institute (Amii)\\
  \texttt{\{matthew.e.taylor,nnissen,wang17\}@ualberta.ca}
  \and
  Neda Navidi\\
  École de technologie supérieure (ETS)\\
  \texttt{neda.navidi@lassena.etsmtl.ca}
}



\date

\maketitle

\begin{abstract}

Reinforcement learning (RL) is a popular machine learning paradigm for game playing, robotics control, and other sequential decision tasks. However, RL agents often have long learning times with high data requirements because they begin by acting randomly. In order to better learn in complex tasks, this article argues that an external teacher can often significantly help the RL agent learn. 

OpenAI Gym is a common framework for RL research, including a large number of standard environments and agents, making RL research significantly more accessible. This article introduces our new open-source RL framework, the {\sc Human Input Parsing Platform for Openai Gym} (HIPPO Gym), and the design decisions that went into its creation. The goal of this platform is to facilitate human-RL research, again lowering the bar so that more researchers can quickly investigate different ways that human teachers could assist RL agents, including learning from demonstrations, learning from feedback, or curriculum learning.

\end{abstract}

\section{Introduction}
Reinforcement learning (RL) is a type of machine learning that lets virtual or physical agents learn through experience, often finding novel solutions to difficult problems and exceeding human performance. 
RL has had many exciting successes, including video game playing~\cite{mnih2015humanlevel}, robotics~\cite{DBLP:conf/corl/MahmoodKVMB18}, stock market trading~\cite{Aiden}, and data center optimization~\cite{Datacenter}. Unfortunately, there are still relatively few real-world, deployed, RL success stories.  One reason is that learning a policy can be very sample inefficient (e.g., Open AI Five used 180 years of simulated training data per day via massive parallelism on many servers~\cite{OpenAIFive}). One reason for this is that RL has traditionally focused on how agents can learn from the ground up.\footnote{For example, Sutton and Barto's influential RL textbook~\cite{sutton2018reinforcement} does not mention human interaction, transfer learning, etc., eschewing external knowledge sources.} While such research is absolutely important, we argue that we need to also better allow RL agents to learn from others, whether programs, agents, or humans.\footnote{RL agents could theoretically treat all input as sensory input, considering it part of the environment. However, it is more practical to program in the ability to leverage advice, rather than requiring the agent to learn the about the special significance, and interpretation, of advice.}
Cheating should be encouraged!\footnote{Of course, there are also good reasons not to include external information. For instance, it may be much more fruitful to spend time and resources developing better algorithms that can directly benefit from Moore's law and its analogue to computational improvements \cite{Bitter}. While this may indeed be a better approach in the long run, we argue that including these kinds of biases can help agents solve difficult RL problems today, without waiting for more powerful algorithms that have yet to be developed.}


Rather than the typical RL setting, in order to reduce the potentially substantial costs of environmental interactions and compute (in terms of time, money, wear and tear on the robot, etc.) and to provide better initial performance, we consider how an RL \emph{student} can receive help from a \emph{teacher} via additional guidance. Our long-term interest in such research is to enable RL to be successfully deployed in more real-world scenarios by focusing exploration and jumpstarting initial behavior to quickly reach high quality policies. For example, our Human-Agent Transfer (HAT) algorithm~\cite{HAT} used 3 minutes of human guidance to save 7 hours of agent learning time in a simulated soccer environment. While such initial successes like HAT are encouraging, many questions must be addressed before these techniques can reliably improve RL performance. Such approaches will allow existing programs and humans to provide guidance to an RL agent, significantly improving RL algorithms so that they can learn better performance faster, relative to 1) learning without external guidance and 2) existing human/agent guidance algorithms.

This brief article has two goals. First, to highlight open and exciting problems, relative to existing work, and to encourage additional research in this area. Second, to introduce an open-source software platform that enables human-in-the-loop RL experiments to easily scale to hundreds or thousands of users.

\begin{figure}[t]
\centering
\includegraphics[width=0.5\textwidth]{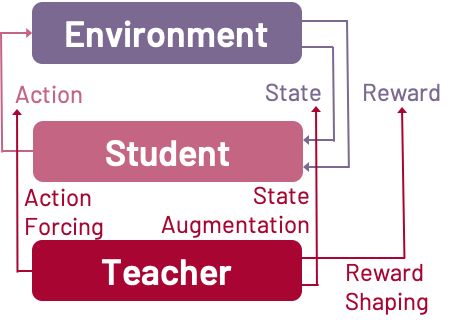}
\caption{\label{fig:studentteacher}A student agent can both learn by interacting with the environment, and directly or indirectly from a teacher. Examples of assistance include suggesting or forcing the student to execute an action, adding additional information to the student's state, and creating a more informative reward signal.} 
\end{figure}

\section{Reinforcement Learning Background}

Reinforcement learning considers the problem of how an agent should act in an environment over time to maximize a reward signal (in expectation). We can formalize the interaction of an agent with its environment as a Markov Decision Process (MDP).

An MDP $M$ is a 5-tuple $(\mathcal{S}, \mathcal{A}, p, r, \gamma)$, where $\mathcal{S}$ is the set of states in the environment, $\mathcal{A}$ is the set of actions the agent can execute, $p(s' | s, a)$ is the transition function that gives the probability of reaching state $s'$ from $s$ after taking action $a$, $r(s)$ is the reward function that gives the immediate reward for reaching state $s$, and $\gamma$ is a $ (0,1]$ discount factor.


At each discrete time step $t$, the agent uses its current state to select an action according to its \emph{policy} $\pi(s)$. The goal is to learn to approach or reach an optimal policy,  $\pi^\star$, which maximizes the expected discounted sum of rewards from now until the end of an episode at time $T$ (or $\infty$ in the non-episodic case):
\[
\mathbb{E} \left[ \sum_{i=t}^{T} \gamma^{i-t} r(s_t) \right]
\]

There are many ways to try to learn to reach, or approximate, $\pi^\star$, including model-free methods that learn how to act in the environment and model-learning methods that can incorporate planning. One common approach is to not learn $\pi$ directly, but to instead learn an action-value function that estimates how good a given action will be in some state when following the current policy: 
\[
q_\pi(s_t, a_t) = \sum_{s_{t+1}, r_t} p(s_{t+1}, r_t | s_t, a_t) \left[ r_t + \gamma \max_{a_{t+1}} q_\pi(s_{t+1}, a_{t+1}) \right]
\]
Eventually, q-values should converge towards $q_{\pi^\star}$, at which point the agent would have learned the optimal policy, $\pi^\star$.

Most real-world domains do not have only tabular states or actions --- to learn in continuous states spaces and/or with continuous action spaces, some type of function approximation is necessary. Currently (deep) neural networks are often used. For an introduction to such learning methods, please see~\cite{sutton2018reinforcement}.

\section{Current Speedup Approaches}

There are many existing approaches leveraging knowledge --- for instance, even using batch or offline RL can be considered ``existing knowledge.'' For an overview, please see our recent survey~\cite{bignold20}. In this article, we specifically focus on a student-teacher framework, where the teacher can be a human, an RL agent, or a program, and the student is an RL agent.

The goals for such an approach can be to
\begin{enumerate}
    \item allow an RL student to improve its learning performance (relative to learning without guidance from a teacher); 
    \item ensure that the student's final performance is not harmed by a suboptimal teacher;
    \item minimize the cognitive load or stress on the human teacher;
    \item minimize the amount of advice needed from the teacher; and
    \item make the best use of whatever advice is provided by the teacher.
\end{enumerate}

A teacher could proactively provide advice~\cite{2014connectionscience-taylor} because it knows better than the student, a student could ask for advice~\cite{da2020uncertainty} because it knows when it is confused, or a combination of both could happen simultaneously~\cite{Ofra16,agentsTeachingAgents}. 
Furthermore, the advice could be provided up front (e.g., a human records a number of demonstrations for an agent~\cite{LfD}) or could be provided over time (e.g., a demonstrator could provide new labels over time~\cite{Dagger}).
The guidance could be free or costly, and unlimited or finite. 

This section briefly reviews a selection of existing approaches where an RL student can improve its learning performance. Generally, we assume a teacher's goals are aligned with the student and its performance is (at least initially) better than the student.

\begin{figure}[t]
\centering
\includegraphics[width=0.75\textwidth]{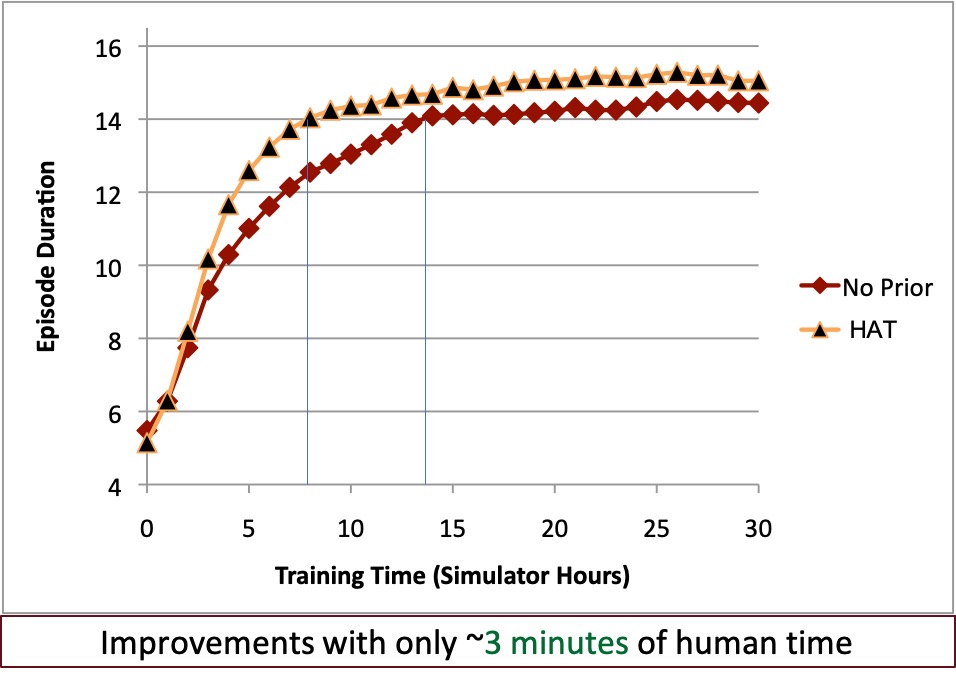}
\caption{\label{fig:HAT}An agent could learn using reinforcement learning alone, or it could leverage just three minutes of human demonstrations to significantly improve the performance, e.g., the time needed to reach a performance of 14, by using the HAT algorithm~\cite{HAT}.}
\end{figure}

\subsection{Programmatic Teachers}

If the teacher is an RL agent, transfer learning~\cite{jmlr09-taylor} can often be used to directly bring the teacher's ``brain'' (e.g., its $q$-values) possibly with some adaptation, into the student. But, in many cases this is infeasible because the teacher's knowledge is not directly accessible or is incompatible with the student's knowledge representation.

More flexible methods like action advising allow a teacher to tell the student what action to perform on the current timestep. As long as the teacher can suggest an action, it does not matter whether the agent uses a neural network, a table, a PID controller, or even a carefully hand-coded policy. The advice could be provided if the student is uncertain and it asks the teacher for help~\cite{da2020uncertainty}, the teacher may proactively provide guidance when it thinks the student is about to make a large mistake~\cite{2014connectionscience-taylor}, or a combination of the two~\cite{Ofra16}. There is typically a fixed teaching budget (e.g., a constraint) or a set teaching cost (e.g., a multi-objective or optimal stopping problem). As long as the student eventually stops receiving advice, the optimal policy is guaranteed not to change~\cite{2016ijcai-zhan}. Existing methods often use heuristics to decide when to request/provide guidance, but a teacher could also learn how to teach~\cite{2017Fachantidis-Taylor-Vlahavas}, or agents could both learn when to ask for, and provide, guidance~\cite{omidshafiei2018learning,JAAMAS20-Leno} simultaneously. 

\subsection{Human Teachers}

In the case of a human teacher, a transfer learning approach that directly uses the teacher's learned policy or q-values, is no longer available. But methods like action advising still apply. Other methods, such as leveraging demonstrations, can be used by programmatic teachers but are even more common with human teachers.
For instance, demonstrations can help initialize a neural network~\cite{gabe_du_taylor_2019,DQFD} that an RL agent uses for representation learning and/or to help learn a policy. Demonstrations could also be used to train a classifier to estimate what action the demonstrator would take in any state~\cite{HAT,2017ijcai-wang}. Another type of guidance is human feedback---the participant could give qualitative feedback (``good robot'' vs. ``bad robot'') that the agent could directly learn from~\cite{TAMER,SABL,COACH}, or the agent could learn from a combination of this feedback and the environmental reward~\cite{PolicyShaping,TAMERRL}. To help the student learn, a human could also provide: a curriculum of tasks of increasing difficulty for the agent to help it learn~\cite{2018curriculadesign}, a shaping reward~\cite{2020ala-paniz,ng1999policy}, natural language advice~\cite{DBLP:conf/ijcai/LuketinaNFFAGWR19}, 
salient regions of the state space~\cite{2020GuanExplanation},
and advice as statements in logic~\cite{Pengo} or code~\cite{rosenfeld_cohen_taylor_kraus_2018}.

\section{Selected Open Problems}

While the previous section gave a brief overview to existing approaches, there are still many open questions. This section highlights questions related to different types of teachers and related to evaluation.

\subsection{Teacher-Dependent Approaches}

This section considers open problems that depend on the type of teacher: both human and programmatic teachers, teachers that are agents, and teachers that are human.

\subsubsection{Human or programmatic teachers} both can allow an RL student to decide when or where it should ask for advice. For instance, assuming advice is not free or infinite, value of information (VOI) estimates could help determine when an agent should ask for advice. This VOI estimate will not only depend on the student's current performance and uncertainty, but also on the estimated quality of the teacher's advice, and the cost for this advice. It is not clear yet when it is better for the student to request guidance (because it knows what it does not know) or better for the teacher to provide proactive guidance (because it understands the task better). 
While this research is easier to conduct with programmatic teachers because data is more plentiful, it is ultimately an even more important for a human teacher, because humans will provide relatively less advice.
 
Existing work (including our own) almost exclusively focuses on demonstrating that a particular method can help improve student learning, rather than understanding where and why different kinds of guidance work best.
For instance, a teacher providing an action suggestion may be particularly useful if the agent has a large action space. In contrast, providing reward feedback may be preferable if the teacher is unable to make fast, low-level action decisions and the environmental reward signal is sparse.

While most current work focuses on a single student and a single teacher, other combinations are possible. For instance, a student could decide which of multiple teachers to query based on their abilities~\cite{li_wei_kudenko_2019} and those teachers could even have different costs. Or teachers could explicitly collaborate, such as forming an ensemble. Similarly, multiple students could have access to a single student, and the students could coordinate among themselves to decide which one asks for guidance~\cite{DBLP:journals/ijsr/ChernovaV10}.

\subsubsection{Agent teachers} must consider when and where to proactively provide advice. In addition to heuristic and learned methods, there could also be advanced modeling methods. In addition to the teacher estimating the student's policy or the student's performance, a deeper level of understanding could allow it to provide more targeted and impactful advice. In the limit, if the teacher fully knew the environment, the student's prior knowledge, and the student's learning algorithm, it could treat teaching as a planning problem~\cite{cdb2010} where it figured out the optimal set of advice to provide to find the most efficient learning outcome.

\subsubsection{Human teachers} introduce many additional challenges. For instance, we now worry about trying to keep them engaged~\cite{DBLP:conf/atal/LiHWK13}, how they might naturally teach~\cite{IJSR12-knox} or prefer to teach, or when one method is perceived to be more difficult to teach (e.g., via the NASA TLX~\cite{NASATLX}).

The participant's background may have a direct impact on the usefulness of their guidance or their comfort providing such guidance. For instance, our prior work~\cite{2019sociology} found a statistically significant (with large effect size) correlation between gender, task framing, and the participant's self-reported interest in a robotic task. We had also found that there was a correlation between teleoperating a UAV with gaming experience~\cite{2015ai_hri-scott}. Other work on robot's learning from demonstration via keyboard interaction~\cite{SuayLfD} showed that non-roboticists interacted significantly differently from roboticists, which affected the performance of some algorithms---we want to make sure our methods work for many different kinds of people, not just researchers in an AI lab! We recommend recruiting and studding interactions with diverse set of participants (e.g., different technology exposure, gender, age, education, video game experience, views on AI, and AI backgrounds). 


We should better understand how the participant's proficiency in a task, the participant's understanding of the task, and whether the agent is explainable or interpretable~\cite{heuillet2020explainability,puiutta2020explainable} affects the quality of the participant's advice, and the resultant quality of the student's learning.

Many studies on human-in-the-loop RL (again, including some of our own), leverage partially trained agents as a stand-in for human advice. However, it is not clear how such automated teachers differ from actual humans. It is likely that there are differences such that a method tuned on automated teachers should be modified for human teachers, and vice versa. Similarly, if we were able to create more realistic teaching agents, the difference between automated teachers and human teachers could be reduced.

Other future studies could also consider questions such as:
\begin{itemize}
    \item How do be best explain to human participants how to provide the most useful feedback for a particular type of learning algorithm?
    \item How do we understand what type of modality a human participant would prefer to use when teaching, and why?
    \item Can we account for biases in how people naturally teach to learn better from them?\footnote{For example, in our past work~\cite{2018curriculadesign}, we biased our curriculum learning algorithm to better take advantage of a pattern we found in human participants. Similarly, in a game theory setting, we know human participants are not perfectly rational, and leveraging theories like quantal response equilibrium~\cite{Quantal} and the anchoring effect~\cite{Anchoring} can help better predict and understand human behavior.}
\end{itemize}

\subsection{Where and How to Evaluate}

This section considers the difference between evaluation methodologies when the teacher is programmatic or is human.

\subsubsection{Evaluation with programmatic teachers} is relatively easy because data is plentiful. A teacher, whether a control method, a hard-coded program, or a RL agent, can run indefinitely in simulation to generate data, helping us understand where and how different kinds of teacher and student approaches do or do not work. Such methods are excellent for producing statistical significance and would apply in real-world settings where such a teacher exists. However, it is not currently clear if and when experiments conducted with programmatic teachers will directly apply to human teachers.

\subsubsection{Evaluation with human teachers} is again more difficult. While there are some guides towards how to start testing human-AI~\cite{10.1145/3290605.3300233} or human-ML~\cite{mathewson2019humancentered} algorithms, few machine learning researchers have extensive human subject study experience. This can require non-trivial ramp-up times for researchers.

Some types of human subject experiments need very specific hardware, such as for visual emotion recognition~\cite{cui2020empathic}, gaze tracking~\cite{saran2020efficiently}, or EEG control~\cite{5509734}. However, we believe that many of these types of human-in-the-loop RL experiments can be conducted over the web with a standard computer. We recently created the {\sc Human Input Parsing Platform for Openai Gym} (HIPPO Gym, Figure~\ref{fig:HIPPO}) project~\cite{HIPPOGym}, which has been released as open source.\footnote{An overview can also be found at \url{https://hippogym.irll.net/}.}
This framework allows easy development and deployment of human subject studies, allowing participants to interact with OpenAI Gym environments like Atari~\cite{ALE} and MuJuCo~\cite{mujoco} over the internet. HIPPO Gym currently supports demonstrations, feedback, mouse clicks (e.g., identifying student mistakes~\cite{2015iui-delacruz}), and task speed adjustment. It can run on standalone servers or be easily integrated into Amazon's Web Service and Amazon's Mechanical Turk. We will continue to develop this codebase to allow for additional kinds of interactions and include implemented learning algorithms for different types of guidance. Our hope is that better standardization among human-in-the-loop RL experiments and experimenters would make this research more accessible and replicable.

By designing and running many human subject studies with hundreds of participants, we will better understand what types of guidance are more or less useful. For instance, a teacher providing an action suggestion may be particularly useful if the agent has a large action space, while providing reward feedback may be preferable if the teacher is unable to make fast, low-level decisions. A related goal is to discover general guidelines about when a human teacher would prefer to provide one type of guidance by asking participants to interact with a student in multiple ways. This approach allows us to not only quantitatively measure the impact of student learning, but also to elicit human teacher preferences (e.g., interviews and a 5-point Likert scale questions) and measure if these correlate with perceived task difficulty (e.g., the NASA TLX~\cite{NASATLX}). 


\section{HIPPO Gym}

We have designed, implemented, tested, and released the HIPPO Gym to meet the \emph{desiderata} discussed in previous sections for conducting experiments with human teachers.

\subsection{Design Principles}

HIPPO Gym is built around three core features: modularity, so that different functions and features can be reused; ease of implementation, so that an RL researcher can quickly get up to speed without a deep understanding of web programming; and inclusion of many examples, so that the researcher can begin quickly understanding the benefits and capabilities of the framework without a large up-front time investment.

\begin{figure}[t]
\centering
\includegraphics[width=0.5\textwidth]{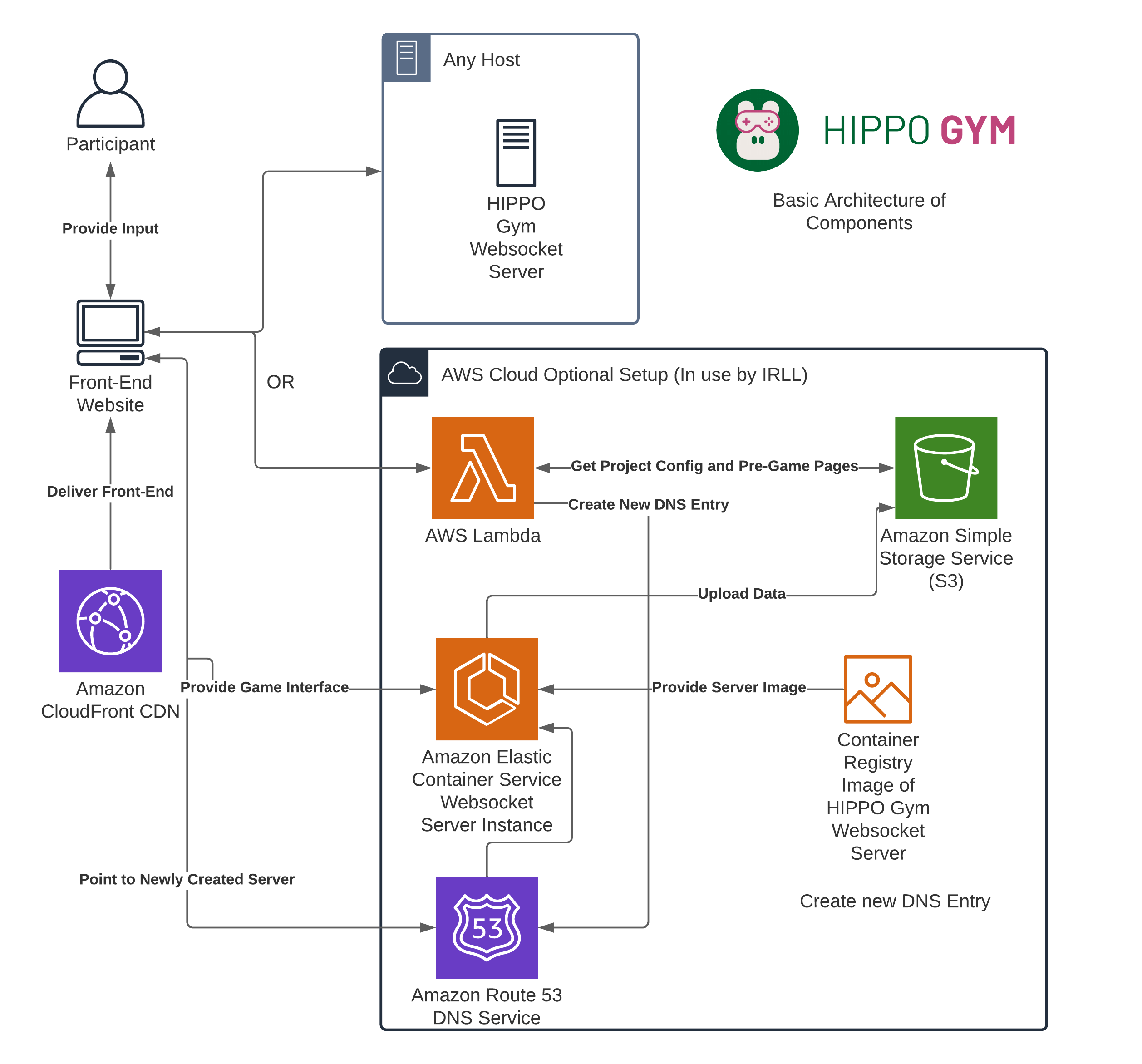}
\caption{\label{fig:HIPPO}Overview of the component structure of HIPPO Gym. Note that the AWS Infrastructure is optional and can be replaced by any server setup.} \end{figure}

HIPPO Gym consists of three primary components: a front-end website for collecting human inputs; a websocket server for connecting the website to an agent and environment; and the cloud infrastructure for deploying experiments. Each component is independent and in practice a researcher may only need to change the websocket server code to integrate their agent/environment in order to run an experiment (see Figure~\ref{fig:HIPPO}).

Ease of use by experimenters (not network engineers) was a critical factor in the development, common languages (python and json) were chosen due to the general familiarity with these tools within the research community. Additionally, the user-facing website is statically deployed and maintained so that any researcher anywhere can use this component with no effort on their part (unless they would prefer to host on their own webserver). Finally, the cloud infrastructure is set up on Amazon Web Services (AWS) so that any researcher with appropriate permissions deploy a research project with only a single command and a configuration file, without requiring significant infrastructure. Instructions and code for using and understanding this infrastructure are available to anyone, so that with a small effort this rapid deployment can be recreated for any group of researchers. However, it is also not mandatory as deployed infrastructure could be as simple as a server sitting on a researcher's desk.

%

\subsection{Code Structure}

This section outlines the structure of the released HIPPO Gym code.

\subsubsection{Front-end Website}

Once human participants are recruited (via email, Amazon's Mechanical Turk, etc.), they can access the front-end user interface.
The IRL Lab hosted version lives at \url{https://irll.net} and is configurable through the use of URL query strings and websocket messages. The functional components of the page (buttons, inputs, and information displays) are determined by the websocket message from the server---a researcher can determine how this web page looks and functions without ever changing its code. Additionally, the project and user IDs, as well as server location, debugging features, and even look and feel via css are controlled via URL query strings. By changing the link to the page, a researcher can control all of these aspects. There is little need for a researcher to do any browser-specific coding.

\begin{figure}[t]
\centering
\includegraphics[width=0.48\textwidth]{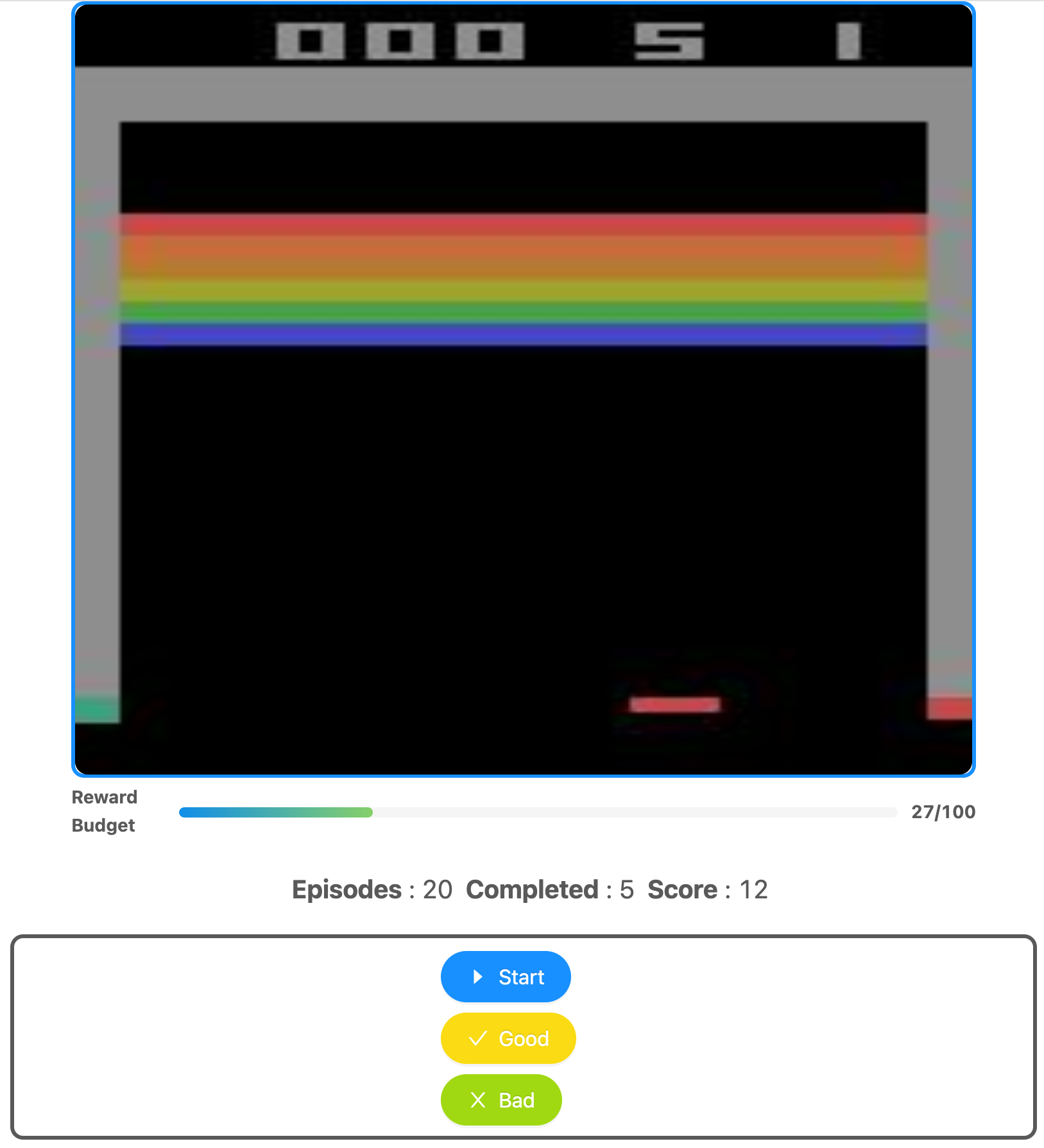}
\includegraphics[width=0.48\textwidth]{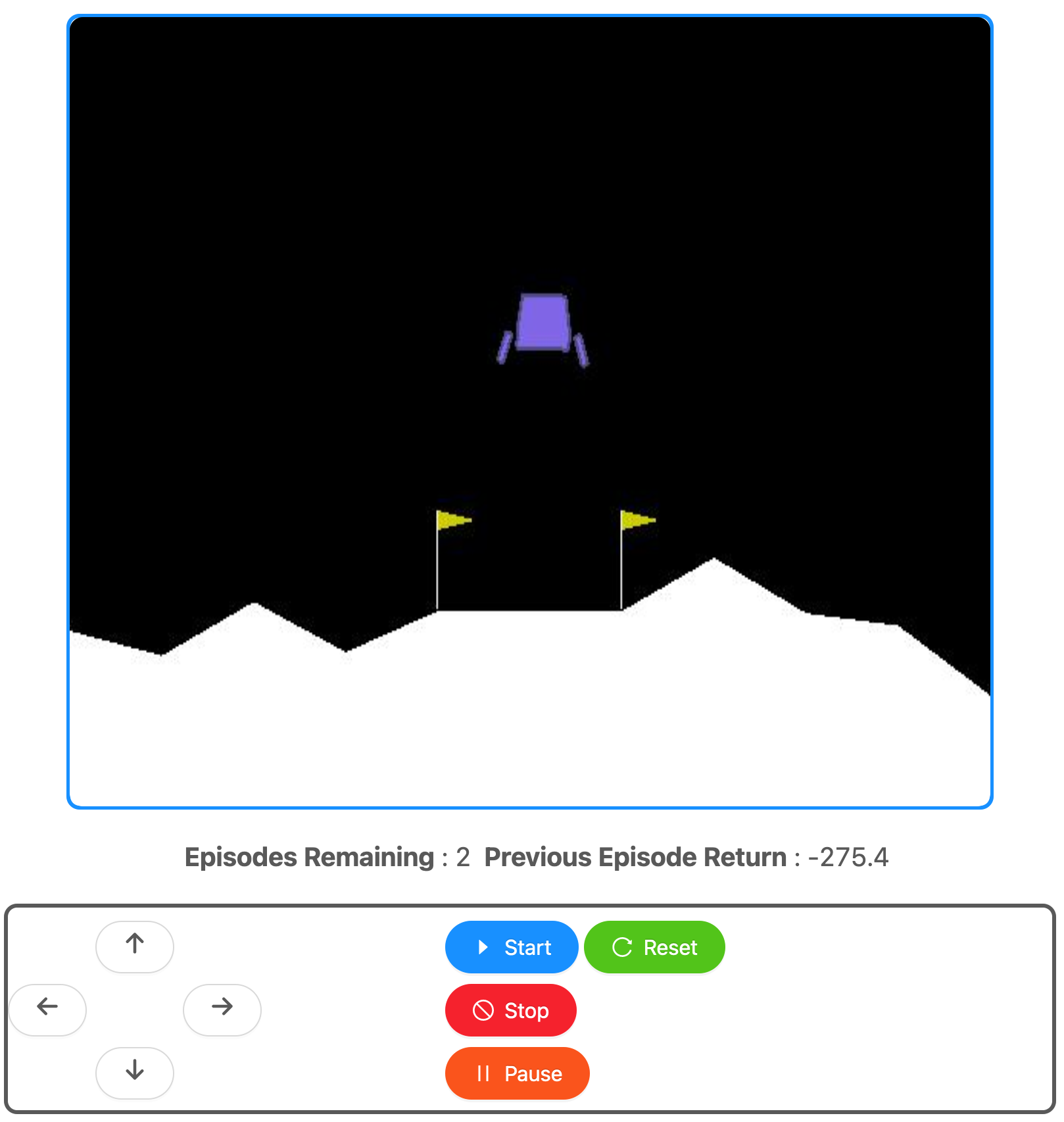}
\caption{\label{fig:HGYM1}In HIPPO Gym, Breakout could be trained with positive and negative human feedback (left) and Lunar Lander could be trained by giving demonstrations (right). } \end{figure}

\begin{figure}[t]
\centering
\includegraphics[width=0.65\textwidth]{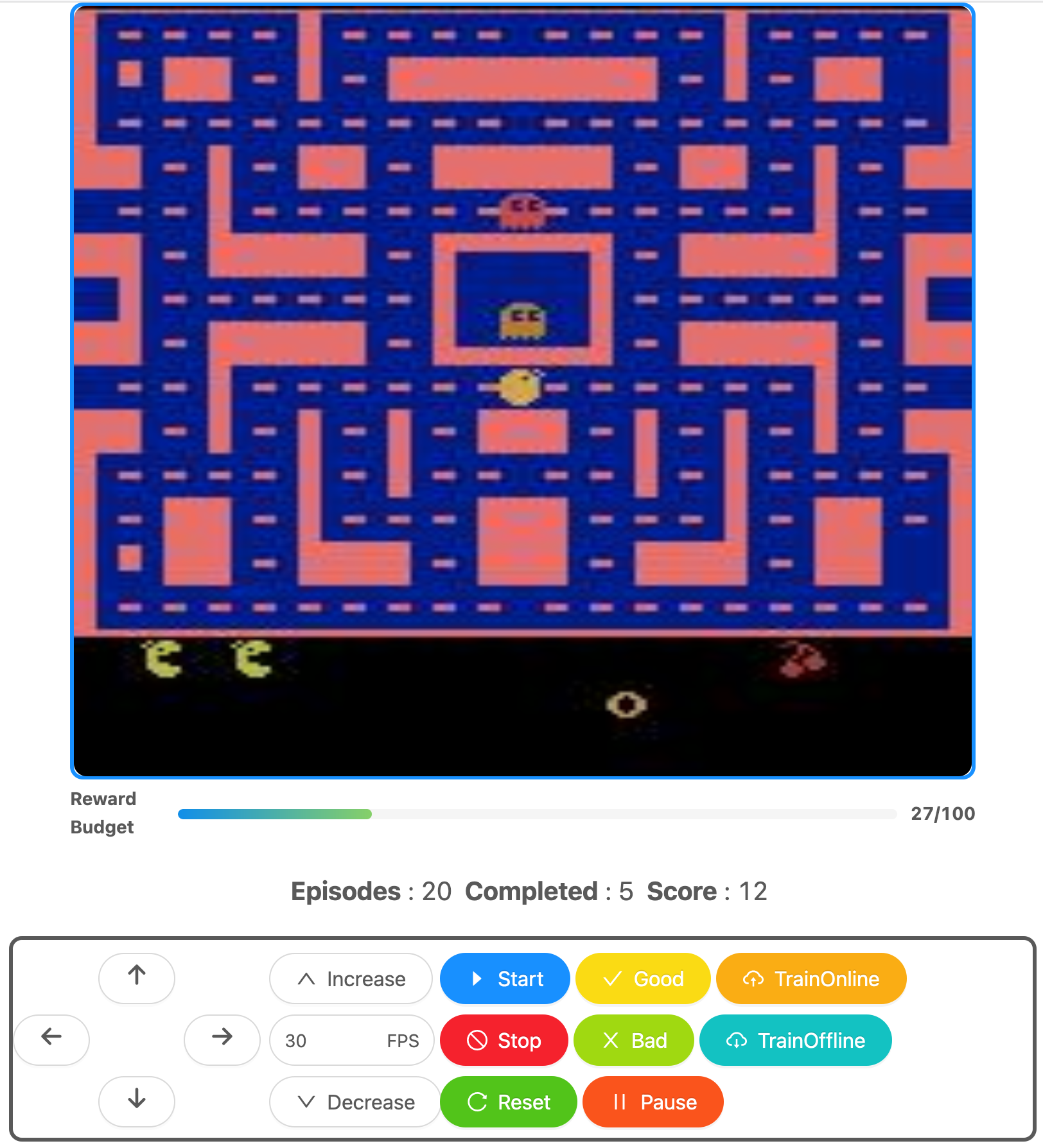}
\caption{\label{fig:HGYM2}This screenshot shows the training Ms.~Pac-Man in HIPPO Gym by using a set number of good/bad feedbacks and demonstrations. Note additional options, which can be easily (de)activated, include increase speed, decrease speed, start, stop, pause, reset, train offline, and train online.} \end{figure}

Examples of the front end for Breakout and Lunar Lander are in Figure~\ref{fig:HGYM1} and Ms.~Pac-Man is shown in Figure~\ref{fig:HGYM2}.

\subsubsection{Back-end Websocket Server}

This layer provides communication between the user interface (UI) in the browser and the agent/environment. At its core, this component sends UI feature instructions to the front end, along with the frame image for the current step in the environment, and then receives and parses input messages from the browser to provide the relevant information to the experimenter's code. There are also functions for saving data and (if appropriate) uploading that data into cloud storage for later use.

The back end is written in python and uses json for websocket messaging. This combination is not the most efficient and in fact limits concurrent users on a single server. However, based on the nature of our targeted research projects, these limitations are far outweighed by the familiarity of researchers with these technologies allowing for much faster development and iteration of experiments.

\subsubsection{Infrastructure}

Infrastructure for any complex project distributed over the web can become quite complex. Given that researcher's time is best spent doing research and not managing infrastructure, we aimed to remove as much infrastructure work as possible. The front-end component is statically deployed on a content delivery network (CDN) and managed by the IRL Lab. Any researcher may use this deployed version for their experiments, removing an entire piece of infrastructure from their process. The front-end code is also open sourced and available through GitHub should anyone want to host their own version or should the IRL Lab hosted version become unavailable.

Server infrastructure becomes more difficult to abstract away. At the base level, HIPPO Gym can be run on any machine, optionally via containerization, by passing either the IP address or the domain name of the machine via query string to the front end. This is especially useful during development and testing of an experiment, and may work in production for small-scale experiments, or experiments that require a very specific hardware configuration. In general, however, this is not the best setup for production deployment of an experiment. Therefore, we created a distributed system of AWS services to provide all of the necessary infrastructure (DNS, on-demand containerized servers, long-term data storage, SSL certificates, etc.) to researchers with a simple configuration file and a single command. This allows our group to deploy and update experiments without touching any infrastructure while still benefiting from all the advantages of AWS cloud services. Should another group or individual wish to replicate the setup for themselves, all the required code and setup instructions are available via GitHub.

Functionally, the infrastructure works as such: the front end is statically hosted with the CloudFront CDN. When a user lands on the page, information is read from the query string of the link that brought them to the page and the correct project is loaded, or the `game' page starts immediately pointing to the given server address. A loaded project may have defined pre-game pages including questionnaires and consent forms. During the pre-game phase the AWS Elastic Container Service is used to start a server for this individual user, the servers are all ephemeral, created when there is a user and destroyed after the user session is complete in order to substantially reduce costs. In order to support SSL, a DNS entry is made in Route53 (the AWS DNS Service), which will point to the just-created server. The front end will then connect to the new server via websocket. The server will pass information about the required UI inputs and they will be loaded along with the first frame of the game. The user then participates in the game in the manner intended by the researcher. When the trial completes the user is shown the next page, often a thank you page, but possibly a redirect elsewhere, further instructions, or even another game page. Once the server is no longer required, the saved data from the trial is uploaded to S3 (AWS Simple Storage Service), which provides an easily accessible long-term file storage, and the server is then destroyed. If a user abandons a trial before completing an experiment, the server has a fail-safe timeout, after which it will be destroyed. This system is extremely cost effective because the infrastructure only has cost when there is a live participant---there is no substantial idle time.

\subsection{Current Abilities}

Currently, HIPPO Gym works well with discrete action space environments from OpenAI Gym including Classic Control and Box2D environments. Continuous action space environments are compatible but require some additional configuration on the part of the researcher.

Available human inputs (all optional) include: all directions, fire, start, pause, stop (end trial), reset (end episode), increase/decrease frame rate, positive feedback (good), negative feedback (bad). It is also possible for a researcher to define other inputs for which they wish to define an action. Users are also able to click on the game window, recording the x and y coordinates, which may be used for identifying points of interest or identifying errors~\cite{2015iui-delacruz}.

Feedback or information can be passed to the user. This information will typically include a score and progress information, but a budget bar is also available that shows how many points of feedback have been given out of a set maximum.

An optional debug mode facilitates development by showing all of the websocket messages in real time.

The base repository for HIPPO Gym includes example integration of both TAMER~\cite{TAMER} and COACH~\cite{COACH} algorithms using tile coding for the Mountain Car environment.

\subsection{Future Enhancements}

In the future, in addition to increased documentation and inevitable bug fixes, we plan the following enhancements:
\begin{itemize}
    \item Implement more OpenAI Gym compatible environments such as Mar.io and Minecraft-type environments.
    \item Allow the user to fast forward and rewind in order to provide feedback.
    \item Provide code supporting additional algorithms that can learn from human teachers (with or without an additional environmental reward) out of the box. 
    \item Introduce more generalized agent examples that are not specific to any particular function approximator.
\end{itemize}

\section{Conclusion}

This article has summarized some current approaches that allow RL students to learn from human or agent teachers, as well as important open questions. It also introduced HIPPO Gym, an open-source framework for human subject studies in human-RL collaboration. Our hope is that by outlining and motivating these questions, and providing a way for researchers to more quickly conduct human experiments, this article will enable a new generation of researchers to begin studying these important questions at the intersection of machine learning and human subjects.

\section{Acknowledgements}
All authors contributed to writing this article.
\begin{itemize}
    \item Taylor conceptualized and directed the project, secured funding, and was the primary author on this paper.
    \item Nissen drove back-end code development, testing, and documentation; lead the integration with Amazon Web Services; and took lead on user testing.
    \item Wang drove the front-end code development, testing, and documentation.
    \item Navidi provided the initial approach for interacting with OpenAI agents over a website, as well as assisted with code reviews.
\end{itemize}

We appreciate help and feedback from other students in the IRL Lab, including
Calarina Muslimani, Rohan Nuttall, and Volodymyr Tkachuk.

\bibliographystyle{spmpsci}
\bibliography{references}

%
%


\end{document}